# REDUPLICATED MWE (RMWE) HELPS IN IMPROVING THE CRF BASED MANIPURI POS TAGGER

Kishorjit Nongmeikapam[1], Lairenlakpam Nonglenjaoba[2], Yumnam Nirmal[3] and Sivaji Bandhyopadhyay[4]

[1, 2]Department of Computer Sc. & Engineering, MIT, Manipur University, Imphal, India
{kishorjit.nongmeikapa,nonglen.ran}@gmail.com
[3]Department of Computer Sc., Manipur University, Imphal, India
yumnamnirmal@gmail.com
[4]Department of Computer Sc. & Engineering, Jadavpur University, Kolkata, India
sivaji_cse_ju@yahoo.com

## ABSTRACT

*This paper gives a detail overview about the modified features selection in CRF (Conditional Random Field) based Manipuri POS (Part of Speech) tagging. Selection of features is so important in CRF that the better are the features then the better are the outputs. This work is an attempt or an experiment to make the previous work more efficient. Multiple new features are tried to run the CRF and again tried with the Reduplicated Multiword Expression (RMWE) as another feature. The CRF run with RMWE because Manipuri is rich of RMWE and identification of RMWE becomes one of the necessities to bring up the result of POS tagging. The new CRF system shows a Recall of **78.22%**, Precision of **73.15%** and F-measure of **75.60%**. With the identification of RMWE and considering it as a feature makes an improvement to a Recall of **80.20%**, Precision of **74.31%** and F-measure of **77.14%**.*

## 1. INTRODUCTION

The Manipuri language or simply Manipuri is among the 21 scheduled language of Indian Constitution. This language is mainly spoken in Manipur in India and in some parts of Myanmar and Bangladesh. Manipuri uses two scripts, one is the borrowed Bengali script and another is the original script that is Meitei Mayek (Script). Our present work adopts the Manipuri in Bengali Script. It is because most of the Manipuri manuscripts are written in Bengali script and thus the corpus collection for Manipuri in Bengali script becomes an easy task.

This language is a Tibeto-Burman language and the distinction from other language Indian is its highly agglutinative nature. As an example (observe Section 2) single word can have 10 suffixes or more. The affixes play the most important role in the structure of the language. The affixes play an important role in determining the POS of a word. Apart from the agglutinative nature, Manipuri is a mono-syllabic, influenced and enriched by the Indo-Aryan languages of Sanskrit origin and English. A clear-cut demarcation between morphology and syntax is not possible in this language. In Manipuri, words are formed in three processes called affixation, derivation and compounding as mentioned in [1]. In Manipuri roots are of two types, they are *free* and *bound* root [2]. The majority of the roots found in the language are bound and the affixes are the determining factor of the class of the words in the language.

The POS tagging is an important topic in the application of Information Retrieval (IR), Question Answering (QA), Summarization, Machine Translation (MT), Event Tracking etc. for all languages and so is for Manipuri. Lack of an efficient POS tagger for Manipuri hampers the other research area of NLP. So the real challenges to bring up the accuracy level motivate us for more experiment and more research in this area.

Every language linguistically has their own POS and computer experts tried their best for the automatic POS tagging using machine. Different attempts are seen for different major languages using different approaches such as for English a Simple Rule-based Part of Speech Tagger is reported in [3], transformation-based error-driven learning [4], decision trees models applied to labelling of texts with parts of speech is reported in [5], Markov model in [6], maximum entropy methods [7] and Part-of-speech tagging using a Hidden Markov Model (HMM) in [8]. For Chinese, the works are found ranging from rule based, HMM to Genetic Algorithm [9]-[12]. For Indian languages like Bengali works are reported in [13]-[15] and for Hindi [16]. Works of POS tagging using CRF can also be seen in [17].

Works of Manipuri POS tagging are reported in [18]-[19]. For the identification of Reduplicated Multiword Expression (RMWE) is reported in [20]. Identification of RMWE using CRF is reported in [21] and improvement of MWE using RMWE is reported in [22]. Web Based Manipuri Corpus for Multiword NER and Reduplicated MWEs Identification using SVM is reported in [23].

Section 2 describes about the agglutinative nature of Manipuri which leads us to the idea of stemming. Section 3 gives the idea of Manipuri Reduplicated MWEs, stemming of Manipuri words so that it can be used as a feature is describe in Section 4. Section 5 gives the concept of Conditional Random Field (CRF). CRF model and Feature selection is described in Section 6 and Section 7 reports the experiments and the evaluation, improvement using reduplicated MWEs is discussed in Section 8 and the conclusion is drawn in Section 9.

## 2. ROOT AND AN EXAMPLE OF AN AGGLUTINATIVE MANIPURI WORD

In Manipuri roots are generally considered by replacing a large number of morphemes. They are the carrier of principal meaning in the composition of a word [2]. Roots are different in different languages. In English roots are almost free which means that there are separate sets of roots that denote separate grammatical categories such as noun (box, boy, bird etc), verb (run, cry, try etc), adjective (tall, thin, large etc) etc. This is not in the case of Manipuri since there are no separate roots for adjective and verb in Manipuri.

In Manipuri roots are of two types, they are *free* and *bound* root. Free roots can stand alone without suffixes in a sentence and bound root takes other affixes excepting the one with free roots.

This language is highly agglutinative and to prove with this point let us site an example word: "পুশিনহনজরমগদবনিদকো" ("pusinhənjərəmgədəbənidəko"), which means "(I wish I) myself would have caused to bring in (the article)". Here there are 10 (ten) suffixes being used in a verbal root, they are "*pu*" is the verbal root which means "to carry", "*sin*"(in or inside), "*hən*" (causative), "*jə*" (reflexive), "*rəm*" (perfective), "*gə*" (associative), "*də*" (particle), "*bə*" (infinitive), "*ni*" (copula), "*də*" (particle) and "*ko*" (endearment or wish).

Altogether 72 (seventy two) affixes are listed in Manipuri out of which 11 (eleven) are prefixes and 61 (sixty one) are suffixes. Table 1 shows the prefixes of 10 (ten number) because the prefix ম (mə) is used as formative and pronomial so only one is included and like the same way table 2 shows the suffixes in Manipuri with only are 55 (fifty five) suffix in the table since some of the suffixes are used with different form of usage such as গুম (gum) which is used as particle as well as proposal negative, দা (də) as particle as well as locative and না (nə) as nominative, adverbial, instrumental or reciprocal.

Table 1. Prefixes in Manipuri.

| Prefixes used in Manipuri |
|---|
| অ, ই, ই, খু, চা, ত, থ, ন, ম and শে |

Table 2. Suffixes in Manipuri.

| Suffixes used in Manipuri |
|---|
| কন, কুম, কো, খরে, খৎ, খাই, খি, খোয়, গা, গনি, গী, গুম, ঙৈ, চা, চো, থ, থৎ, থেক, থোক, দা, দি, দুনা, দে, না, নত্তে, নি, নিং, নু, নে, পী, ফাৎ, বা, বু, মক, মল, মিন, মুক, লে, লা, লক, ল্ম, লি, লী, লূ, লু, লে, লো, লোয়, শনু, শি, শিং, শিন, শু, হৎ and হন |

## 3. REDUPLICATED MULTIWORD EXPRESSIONS (RMWE)

In Manipuri works for identification of reduplicated MWEs has been reported for the first time in [20]. In this paper the process of reduplication is defined as: *'reduplication is that repetition, the result of which constitutes a unit word'*. These single unit words are the RMWE.

The reduplicated MWEs in Manipuri are classified mainly into four different types. These are: 1) Complete Reduplicated MWEs, 2) Mimic Reduplicated MWEs, 3) Echo Reduplicated MWEs and 4) Partial Reduplicated MWEs. Apart from these fours there are also cases of a) Double reduplicated MWEs and b) Semantic Reduplicated MWEs.

### 3.1 Complete Reduplication MWEs

In the complete reduplication MWEs the single word or clause is repeated once forming a single unit regardless of phonological or morphological variations. Interestingly in Manipuri these complete reduplication MWEs can occur as Noun, Adjective, Adverb, *Wh-* question type, Verbs, Command and Request.

#### Noun

Here the reduplication words are generally noun but sometimes the suffixes *–də*, *–gi* and *–ki* are generally added to the second word of reduplication to identify itself as noun. One such MWE is *মরিক মরিক* (*'mərik mərik'*) which means *'drop by drop'*.

Also some of these words are inflected MWEs too with one of the above suffixes, say; *য়ুম য়ুমদা* ( *'yum yum-də'*) which means *'to every house'*. The following example sentence in Manipuri:

*lam lam-gi məsa-gi lon lei*
*place place own-gen language has*

Means, *'Every place/country has its own* language', when translated in English.

#### Adjective

In this type of reduplication MWEs, the first word is repeated to form an adjective by adding *–bə* or *–pə* to the second word. For example, *অনৌ অনৌবা* ( *'ənou əou-ba'*) which means *'new'* and *অটেক অটেকপা* ( *'ətek ətek-pa'* ) which means *'fresh'* etc. The following example sentence in Manipuri:

*əy ətek ətekpa ləy pam-mi*
*I fresh fresh flower like-asp*

Means, *'I like fresh flowers.'* when translated in English.

### Verb

The verb reduplication is different from the reduplication of noun, adjective, adverb and *wh*-question since they are generally of subject-objectless structure. For example, চৎলে চৎলে (*'cətle cətle'*) means *'am going/leaving'*. Sometimes with this reduplication *'I'* can be silent and can mean *'I am going/leaving'*. The verb word generally ends with any of these verbal inflections *-le, -re, -gani, -lurə etc.* depending on the tense.

### Adverb

In case of adverbs *–nə* is usually added to both the words of the reduplication. It is different from the above in such a way that the inflection is added to both the words in reduplication. For example, কপনা কপনা (*'kəp-na kəp-na'*) which means *'cryingly'* and য়ামনা য়ামনা (*'yam-nə yam-nə'*) which means *'very'*. The following example sentence in Manipuri:

*əy kəpna kəpna skul-də cətli*
*I cryingly cryingly school-loc go*

Means, *'I go to school cryingly.'* when translated to English.

### *Wh-* question type

Like English, *wh-* type of questions can be framed in Manipuri but it uses reduplication MWEs. For example:

করি করি (*'kəri kəri'*) means *'what/which'*.
কনা কনা (*'kəna kəna'*) means *'who'*.
কদায় কদায় (*'kəday kəday'*) means *'where'*.
করম করম (*'kərəm kərəm'*) means *'how'*.

The following example sentence in Manipuri:

*kəna kəna lam yengbə cətkani?*
*who who land to see go-asp*

Means *'who will go to see the land?'* when translated to English.

### Command

The reduplication is used in such a manner that the subject is generally a second person but it is not explicitly mentioned in the sentence.

For example, চৎলো চৎলো (*'cətlo cətlo'*) means *'go-comd go-comd'* and which can be translated to English as 'Go.'

### Request

Reduplication can be present in request sentences also.
For example, লেংবিরো লেংবিরো (*'leŋ-bi-ro leŋ-bi-ro'*) means *'please go/please leave'*.

## 3.2 Partial Reduplicated MWEs

In case of partial reduplication the second word carries some part of the first word as an affix to the second word, either as a suffix or a prefix. For example, চৎথোক চৎসিন (*'cət-thok cət-sin'*) means *'to go to and fro'*, শামী লানমী (*'sa-mi lan-mi'*) means *'army'*.

## 3.3 Echo Reduplicated MWEs

The second word does not have a dictionary meaning and is basically an echo word of the first word. For example, থকসি খাসি *('thk-si kha-si')* means *'good manner'*. Here the first word has a dictionary meaning *'good manner'* but the second word does not have a dictionary meaning and is an echo of the first word.

### 3.4 Mimic Reduplicated MWEs

In the mimic reduplication the words are complete reduplication but the morphemes are onomatopoetic, usually emotional or natural sounds. For example, করক করক (*'khrək khrək'*) means *'cracking sound of earth in drought'*.

### 3.5 Double Reduplicated MWEs

In double Reduplicated MWE there consist of three words, where the prefix or suffix of the first two words is reduplicated but in the third word the prefix or suffix is absent. An example of double prefix reduplication is ইমুন ইমুন মুনবা (*'i-mun i-mun mun-ba'*) which means, *'completely ripe'*. It may be noted that the prefix is duplicated in the first two words while in the following example suffix reduplication take place, ঙৌশোক ঙৌশোক ঙৌবা ( *'ŋəω-srok ŋəω-srok ŋəω-ba'*) whichmeans *'shining white'*.

### 3.6 Semantic Reduplicated MWEs

Both the reduplication words have the same meaning as well as the MWE. Such type of MWEs is very special to the Manipuri language. For example, পামবা কৈ (*'pamba kəy'* ) means *'tiger'* and each of the component words means *'tiger'*. Semantic reduplication exists in Manipuri in abundance as such words have been generated from similar words used by seven clans in Manipur during the evolution of the language.

## 4. STEMMING OF MANIPURI WORDS

The Stem words which are considered as feature for running the CRF follows an algorithm mention in [25]. In this algorithm Manipuri words are stemmed by stripping the suffixes in an iterative manner. As mentioned in Section 2 a word is rich with suffixes and prefixes. In order to stem a word an iterative method of stripping is done by using the acceptable list of prefixes (11 numbers) and suffixes (61 numbers) as mentioned in the Table 1 and Table 2 above.

### 4.1 The Algorithm

This stemmer mainly consist of four algorithms the first one is to read the prefixes, the second one is to read the suffixes, the third one is to identify the stem word removing the prefixes and the last algorithm is to identify the stem word removing the suffixes.

Two file, *prefixes_list* and *suffixes_list* are created for prefixes and suffixes of Manipuri. In order to test the system another testing file, *test_file* is used.
The prefixes and suffixes are removed in an iterative approach as shown in the algorithm 3 and algorithm 4 until all the affixes are removed. The stem word is stored in *stemwrd.*

**Algorithm1:** `read_prefixes()`

1. Repeat 2 to 4 until all prefixes ($p_i$) are read from *prefixes_list*
2. Read a *prefix* $p_i$
3. *p_array[i]*=$p_i$
4. *p_wrd_count =i++;*
5. exit

**Algorithm2:** `read_suffixes()`

1. Repeat 2 to 4 until all suffixes ($s_i$) are read from *suffixes_list*
2. Read a suffix $s_i$
3. *s_array[i]*=$s_i$
4. *s_wrd_count=i++;*
5. *exit*

**Algorithm3:** *Stem_removing_prefixes(p_array, p_wrd_count)*

```
1. Repeat 2 to 16 for every word (wi) are read from the
   test_file
2. String stemwrd=" ";
3. for(int j=0;j<p_wrd_count;j++)
4. {
5. if(wi.startsWith(p_array[j]))
6.  {
7.  stemwrd=wi.substring(wi.length()-((wi.length()-
   ((p_array[j].toString()).length()))),wi.length());
8.   wi=stemwrd;
9.  j=-1;
10.       }
11. else
12.       {
13.        stemwrd=wi;
14.       }
15. }
16. write stemwrd;
17. exit;
```

**Algorithm4:** *Stem_removing_suffixes(s_array,s_wrd_count)*

```
1. Repeat 2 to 16 for every word (wi) are read from the
   test_file
2. String stemwrd=" ";
3. for(int j=0;j<s_wrd_count;j++)
4.  {
5.   if(wi.endsWith(s_array[j]))
6.   {
7.   stemwrd=wi.substring(0,wi.indexOf(s_array[j]));
8.   wi=stemwrd;
9.   j=-1;
10.  }
11.  else
12.  {
13.   stemwrd=wi;
14.  }
15. }
16. write stemwrd;
17. exit
```

## 5. CONCEPTS OF CRF

The concept of Conditional Random Field in [24] is developed in order to calculate the conditional probabilities of values on other designated input nodes of undirected graphical models. CRF encodes a conditional probability distribution with a given set of features. It's an unsupervised approach where the system learns by giving some training and can be used for testing other text.

The conditional probability of a state sequence $Y=(y_1, y_2,..y_T)$ given an observation sequence $X=(x_1, x_2,..x_T)$ is calculated as :

$$P(Y|X) = \frac{1}{Z_X} \exp\left(\sum_{t=1}^{T}\sum_{k} \lambda_k f_k(y_{t-1}, y_t, X, t)\right) ----(1)$$

Where, $f_k(y_{t-1}, y_t, X, t)$ is a feature function whose weight $\lambda_k$ is a learnt weight associated with $f_k$ and to be learned via training. The values of the feature functions may range between $-\infty \ldots +\infty$, but typically they are binary. $Z_X$ is normalization factor:

$$Z_X = \sum_y \exp \sum_{t=1}^{T} \sum_k \lambda_k f_k(y_{t-1}, y_t, X, t)) \quad \text{----(2)}$$

Which is calculated in order to makes the probability of all state sequences sum to 1. This is calculated as in HMM and can be obtain efficiently by dynamic programming. Since CRF defines the conditional probability P(Y|X), the appropriate objective for parameter learning is to maximize the conditional likelihood of the state sequence or training data.

$$\sum_{i=1}^{N} \log P(y^i \mid x^i) \quad \text{---(3)}$$

Where, $\{(x^i, y^i)\}$ is the labeled training data.

Gaussian prior on the $\lambda$'s is used to regularize the training (i.e smoothing). If $\lambda \sim N(0,\rho^2)$, the objective becomes,

$$\sum_{i=1}^{N} \log P(y^i \mid x^i) - \sum_k \frac{\lambda_i^2}{2\rho^2} \quad \text{---(4)}$$

The objective is concave, so the *λ's* have a unique set of optimal values.

# 6. CRF MODEL AND FEATURE SELECTION

## 6.1 The CRF Model

The work of [19] also shows the use of CRF in order to tag the POS in a running text. It was the first attempt for POS tagging using CRF and had a low efficiency. This work is an attempt or an experiment to make the previous work more efficient. The C++ based CRF++ 0.53 package[1] is used in this work and it is readily available as open source for segmenting or labeling sequential data.

The CRF model for Manipuri POS tagging (Figure 1) consists of mainly data training and data testing. The important processes required in POS tagging using CRF are feature selection, preprocessing which includes arrangement of tokens or words into sentences with other notations, creation of model file after training and finally the testing with the test corpus.

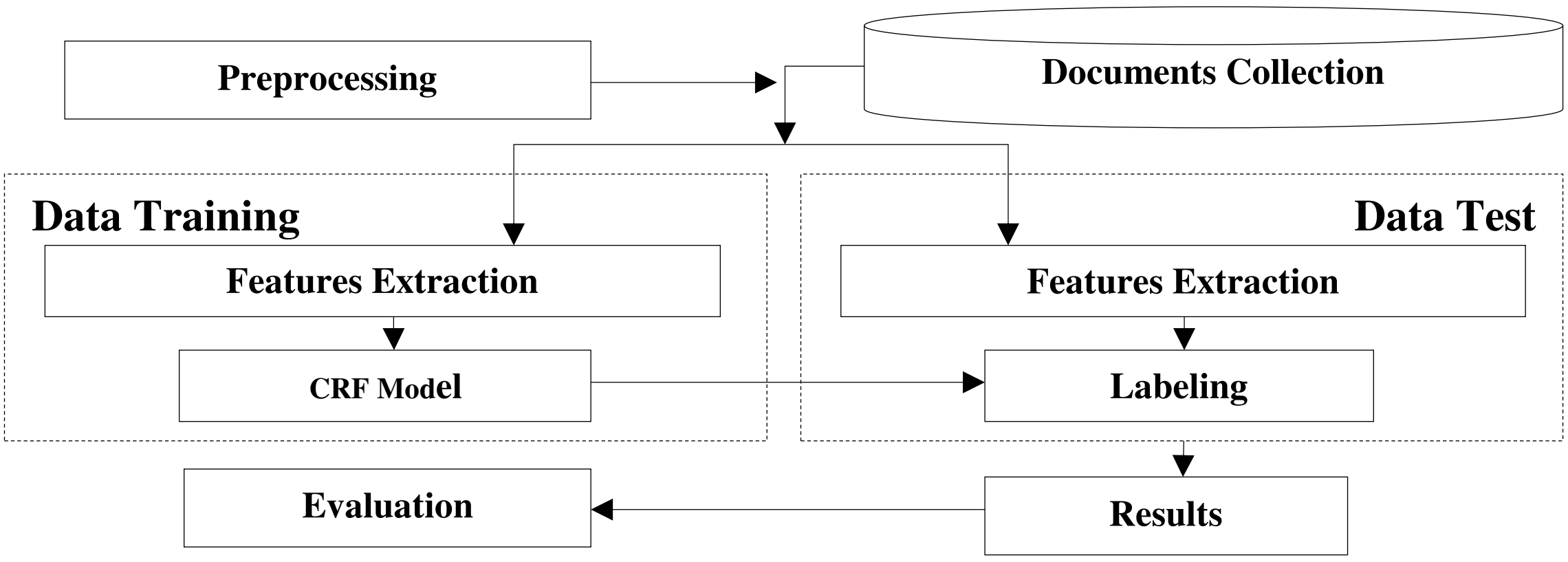

Figure 1.CRF Model of Manipuri POS Tagging

Following sub sections explain the overall process in detail:

[1] http://crfpp.sourceforge.net/

### 6.1 Feature Selection

The feature selection is important in CRF. The various features used in the system are,

**F= { $W_{i-m}$, … ,$W_{i-1}$, $W_i$, $W_{i+1}$, ..., $W_{i+n}$, $SW_{i-m}$, …, $SW_{i-1}$, $SW_i$, $SW_{i+1}$,… , $SW_{i-n}$ , number of acceptable standard suffixes, number of acceptable standard prefixes, acceptable suffixes present in the word, acceptable prefixes present in the word, word length, word frequency, digit feature, symbol feature }**

The details of the set of features that have been applied for POS tagging in Manipuri are as follows:

**Surrounding words as feature:** Preceeding word(s) or the successive word(s) are important in POS tagging because these words play an important role in determining the POS of the present word.

**Surrounding Stem words as feature:** The Stemming algorithm mentioned in Section 4 is used so that the preceding and following stem words of a particular word can be used as features. It is because the preceding and following words influence the present word POS tagging.

**Number of acceptable standard suffixes as feature:** As mention in Section 2, Manipuri being an aggllutinative language the suffixes plays an important in determining the POS of a word. For every word the number of suffixes are identified during stemming and the number of suffixes is used as a feature.

**Number of acceptable standard prefixes as feature:** Same is the case for the prefixes thus it also plays an important role too for Manipuri since it is a highly agglutinative language. For every word the number of prefixes are identified during stemming and the number of prefixes is used as a feature.

**Acceptable suffixes present as feature:** The standard 61 suffixes of Manipuri which are identified is used as one feature. As mention with an example in Section 2, suffixes are appended one after another and so far the maximum number of appended suffixes is reported as ten. So taking into account of such case, for every word ten columns separated by a space are created for every suffix present in the word. A “0” notation is being used in those columns when the word consists of no acceptable suffixes.

**Acceptable prefixes present as feature:** 11 prefixes have been manually identified in Manipuri and the list of prefixes is used as one feature. For every word if the prefix is presents then a column is created mentioning the prefix, otherwise the “0” notation is used. Upto three prefixes are considered for observation.

**Length of the word:** Length of the word is set to 1 if it is greater than 3 otherwise, it is set to 0. Very short words are generally pronouns and rarely proper nouns.

**Word frequency:** A range of frequency for words in the training corpus is set: those words with frequency <100 occurrences are set the value 0, those words which occurs >=100 are set to 1. The word frequency is considered as one feature since occurrence of determiners, conjunctions and pronouns are abundant.

**Digit features:** Quantity measurement, date and monetary values are generally digits. Thus the digit feature is an important feature. A binary notation of ‘1’ is used if the word consist of a digit else ‘0’.

**Symbol feature:** Symbols like $,%, - etc. are meaningful in textual uses, so the feature is set to 1 if it is found in the token otherwise, it is set to 0. This helps to recognize SYM (Symbols) and QFNUM (Quantifier number) tags.

### 6.2 Pre-processing and feature selection

A Manipuri text document is used as an input file. The training and test files consist of multiple tokens. In addition, each token consists of multiple (but fixed number) columns where the columns are used by a template file. The template file gives the complete idea about the feature selection. Each token must be represented in one line, with the columns separated by a white spaces (spaces or tabular characters). A sequence of tokens becomes a **sentence**. Before undergoing training and testing in the CRF the input document is converted into a multiple token file with fixed columns and the template file allows the feature combination and selection which is specified in section 6.1.

Two standard files of multiple tokens with fixed columns are created: one for training and another one for testing. In the training file the last column is manually tagged with all those identified POS tag[2] whereas in the test file we can either use the same tagging for comparisons or only 'O' for all the tokens regardless of POS.

### 6.3 Training to get the Model File

Training of CRF system is done in order to train the system which gives an ouput as a **model file**. The system is feeded with the gold standard training file.

In the training of the CRF we used a **template file** whose function is to choose the features from the feature list. Model file is the output file after the training. Model files are the learnt file by the CRF system. The model file is obtain after training the CRF using the training file. This model file is a ready-made file by the CRF tool for use in the testing process. There is no need of using the template file and training file again since the model file consists of the detail information of the template file and training file.

### 6.4 Testing

The testing proceeds using the model file that have been generated while training the CRF system. One among the two gold standard file which are being created before running the CRF system that is the testing file is used for testing. As mentioned earlier in section 6.2 this file has to be created in the same format as that of training file, i.e., of fixed number of columns with the same field as that of training file.

Thus after testing process the output file is a new file with an extra column which is tagged with the POS tags.

## 7. Experiment and the Result

Manual tagging is time consuming so as mention earlier in Section 6.2 three linguist experts from Linguistic Department, Manipur University is hired for rectifying the spelling and syntax of a sentence. Also they have performed the most important part of POS tagging on the corpus which consists of 25,000 tokens. It is considered to be the Gold standard but later split into two file one for training file and another for testing.

As mention in section 6 a Java program is written in order to identify the required features except the POS tagging since it's manually tag and arrange the feature column wise separated by a space. So, each line becomes a sentence. The last column of both the training and testing files are tagged with the POS. The last column of the test file is also tagged with POS so that the output can be compared easily.

---

[2] http://shiva.iiit.ac.in/SPSAL2007/iiit_tagset_guidelines.pdf

In order to evaluation the experiment, the system used the parameters of Recall, Precision and F-score. It is defined as follows:

Recall, $\mathbf{R} = \dfrac{\textit{No of correct ans given by the system}}{\textit{No of correct ans in the text}}$

Precision, $\mathbf{P} = \dfrac{\textit{No of correct ans given by the system}}{\textit{No of ans given by the system}}$

F-score, $\mathbf{F} = \dfrac{(\beta^2 + 1)\ PR}{\beta^2 P + R}$

Where $\beta$ is one, precision and recall are given equal weight.

Generally a number of problems have been identified because of the typical nature of the Manipuri language. The first thing is the highly agglutinative nature. Another problem is the word category which is not so distinct. The verbs consist of both free and bound category but classifying the bounded categories is a problem. Another problem is to classify basic root forms according to the word class. Although the distinction between the noun class and verb classes is relatively clear; the distinction between nouns and adjectives is often vague. Distinction between a noun and an adverb becomes unclear because structurally a word may be a noun but contextually it is adverb. Also a part of root may also be a prefix, which leads to wrong tagging. The verb morphology is more complex than that of noun. Sometimes two words get fused to form a complete word

## 7.1 Experiment for selection of best feature

As mention in Section 6.2 the preprocessing of the file is done and a total of 25,000 words are divided into 2 file, one consisting of 20000 words and the second file consist of 5000 words. The first file is used for training and the second file is used for testing. As mention earlier, a Java program is used to separate the sentence into equal numbers of columns separated by a blank space with different features.

The experiment is performed with different combinations of features. The features are manually selected in such a way that the result shows an improvement in the F-measure. Among the different experiments with different combinations table 4 list some of the best combinations. Note that the combinations in the list are not the only combinations tried but are some of the best combinations. Table 3 explains the notations used in table 4.

Table 3: Meaning of the notations

| Notation | Meaning |
|---|---|
| **W[-*i*,+*j*]** | Words spanning from the $i^{th}$ left position to the $j^{th}$ right position |
| **SW[-*i*, +*j*]** | Stem words spanning from the $i^{th}$ left to the $j^{th}$ right positions |
| **P[*i*]** | The *i* is the number of acceptable prefixes considered |
| **S[*i*]** | The *i* is the number of acceptable suffixes considered |
| **L** | Word length |
| **F** | Word frequency |
| **NS** | Number of acceptable suffixes |
| **NP** | Number of acceptable prefixes |
| **D** | Digit feature (0 or 1) |
| **SF** | Symbol feature (0 or 1) |

Table 4: Meaning of the notations

| Feature | R (in %) | P (in %) | FS (in %) |
|---|---|---|---|
| W[-2,+1], SW[-1,+1], P[1], S[4], L, F, NS, NP, D, SF | **78.22** | **73.15** | **75.60** |
| W[-2,+2], SW[-2,+1], P[1], S[4], L, F, NS, NP, D, SF | 77.23 | 72.22 | 74.64 |
| W[-2,+3], SW[-2,+2], P[1], S[4], L, F, NS, NP, D, SF | 75.24 | 69.09 | 72.04 |
| W[-3,+1], SW[-3,+1], P[1], S[4], L, F, NS, NP, D, SF | 72.01 | 65.45 | 68.57 |
| W[-3,+3], SW[-3,+2], P[1], S[5], L, F, NS, NP, D | 61.76 | 49.61 | 55.02 |
| W[-3,+4], SW[-2,+3], P[2], S[5], L, F, NS, SF | 53.47 | 42.52 | 47.37 |
| W[-4,+1], SW[-4,+1], P[2], S[6], L, NP, D, SF | 47.01 | 37.60 | 41.78 |
| W[-4,+3], SW[-3,+3], P[3], S[9], L, F, D, SF | 38.00 | 31.93 | 34.70 |
| W[-4,+4], SW[-4,+4], P[3], S[10], NS, NP | 34.65 | 31.82 | 33.17 |

### 7.2 Evaluation and best Feature

The drawback of manual selection is the feature combination. Also the rigidness in the CRF model is the feature combination and feature selection. The best feature so far reported in the previous CRF based model [19] is as follows:

**F= {$W_{i-2}$, $W_{i-1}$, $W_i$, $W_{i+1}$, |prefix|<=n, |suffix|<=n, Dynamic POS tag of the previous word, Digit information, Length of the word, Frequent word, Symbol feature}**

The list consists of surrounding words, prefixes, suffixes, surrounding POS, digit information, word length, word frequency and symbol features. The prefixes and suffixes used are just a combination of the n characters. It does not adopt the standard prefixes and suffixes. The model which has been adopted here has a different list and the best feature is chosen after all possible combinations. The best result is the one which shows the best efficiency among the results. This happens with the following feature:

**F= { $W_{i-2}$, $W_{i-1}$, $W_i$, $W_{i+1}$, $SW_{i-1}$, $SW_i$, $SW_{i+1}$, number of acceptable standard suffixes, number of acceptable standard prefixes, acceptable suffixes present in the word, acceptable prefixes present in the word, word length, word frequency, digit feature, symbol feature }**

The best feature set in the model gives the Recall (**R**) of **78.22%**, Precision (**P**) of **73.15%** and F-measure (**F**) of **75.60%**. The earlier model in [19] reports that the CRF based system shows **72.04%** efficient.

## 8. IMPROVEMENT USING RMWE

With the intention of improving the efficiency of the POS tagging in the CRF system the concept of identifying the RMWE and including it as a feature is tried.

### 8.1. The RMWE identification Model

The Algorithms and models for finding reduplicated MWEs in Manipuri text as suggested in [20] is used for the identification of reduplicated MWEs. The first model is used to identify the complete, mimic, partial, double and echo reduplicated MWEs. Figure 2gives the idea about the model 1.

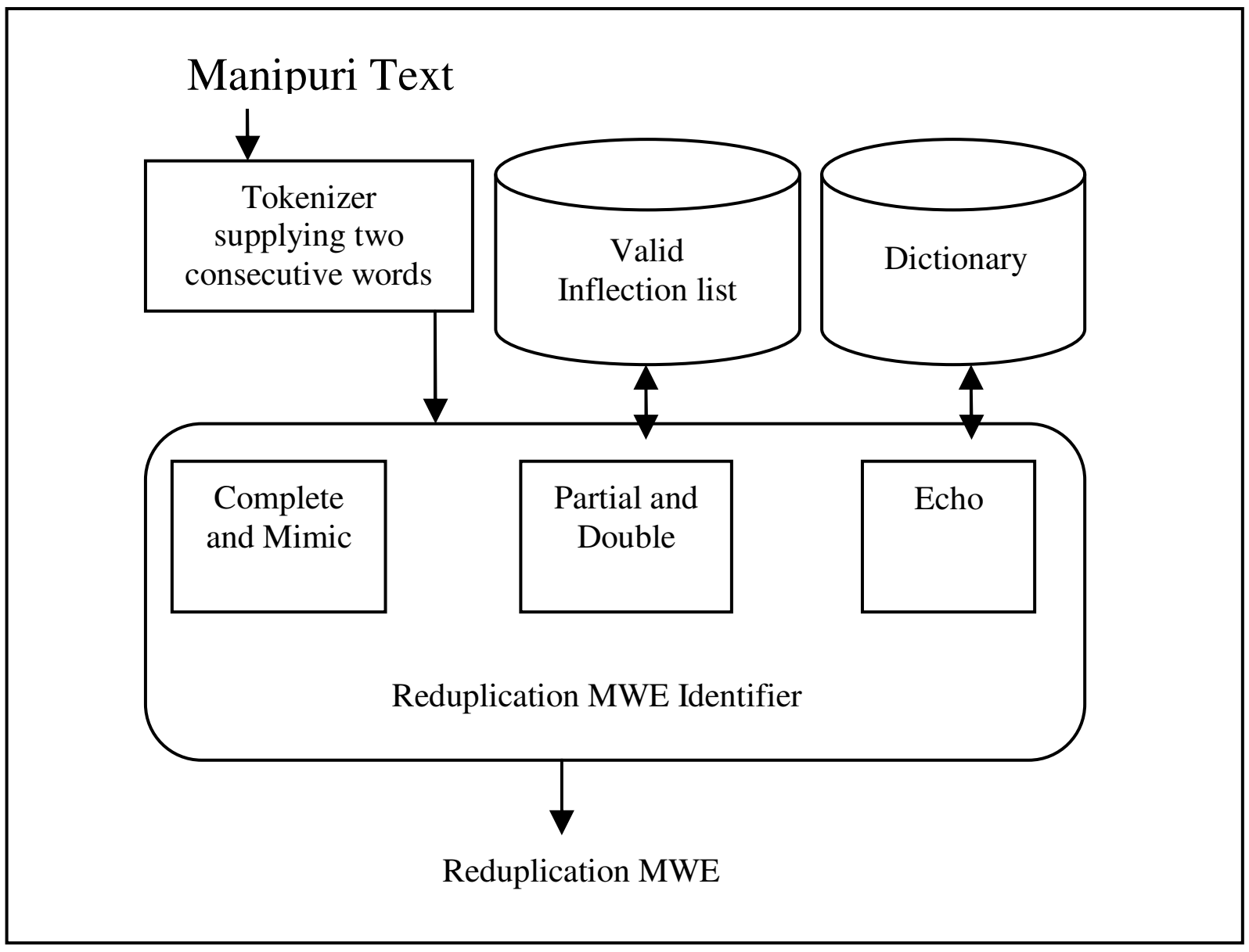

Figure 2: The First model for identifying four types of reduplication MWEs and double reduplication.

The functions performed by the different parts of the proposed model are:

*a)* **Tokenizer,** separates the words based on blank space or special symbols to identify two consecutive words $W_i$ and $W_{i+1}$.

*b)* **Reduplication MWE Identifier,** it verifies the valid inflections present in the words and also checks the semantics of $W_{i+1}$ in the dictionary for the Echo words.

*c)* **Valid Inflection List,** list of commonly used valid inflection is listed. The inflection list is an important resource for MWE identification.

*d)* **Dictionary,** it includes the lexicon and the associated semantics.

## 8.1. The Algorithm for first model

Algorithm to identify the types of reduplicating MWEs:

1. *Repeat 2 to 21 until all the tokens ($W_i$) are read in the text, where (i=1 to n).*
2. *Check whether $W_i$ and $W_{i+1}$ are same word, if same go to 3 else go to 14*
3. *Check whether $W_i$ $W_{i+1}$ is in the dictionary if found then identify as* ***mimic reduplication***
4. *Repeat 5 to 8 until the entire prefixes ($P_k$) list is read, where (k=1 to m)*
5. *temp1= ($W_i$ ) - $P_k$*
6. *Check whether temp1 is a starting substring of $W_{i+2}$ if so go to 9*
7. *temp2= ($W_i$ ) - $S_j$*
8. *Check whether temp2 is a starting substring of $W_{i+2}$ if so go to 9 else go to 13*
9. *Identify it as a* ***double reduplication***
10. *Repeat 11 to 12 until the entire suffixes ($S_j$) list is read, where (j=1 to l).*
11. *temp3=( $W_i$ ) + $S_j$*
12. *Check temp3 and $W_{i+1}$ are same words if same go to 13 else go to 14*
13. *Identify as* ***complete reduplication****.*
14. *Repeat 15 until the entire suffixes ($S_j$) list is read, where (j=1 to l).*
15. *Check $S_j$ is a substring present at the end of $W_i$ and $W_{i+1}$ if so go to 16*
16. *Check whether $W_{i+1}$ is in the dictionary if not identify as* ***echo reduplication*** *else go to 21*
17. *Repeat 18 until the entire prefixes ($P_k$) list is read, where (k=1 to m)*
18. *Check $P_k$ is a substring present at the beginning of $W_i$ and $W_{i+1}$ if so go to 21*
19. *Repeat 20 until the entire prefixes ($P_k$) list are read, where (k=1 to m)*
20. *Check $P_k$ is a substring present at the end of $W_i$ and at the beginning of $W_{i+1}$ if so go to 21*

21. *Identify as* ***partial reduplication****.*
22. *End.*

The second model (Figure 3) is used for identification of semantic reduplicated MWEs.

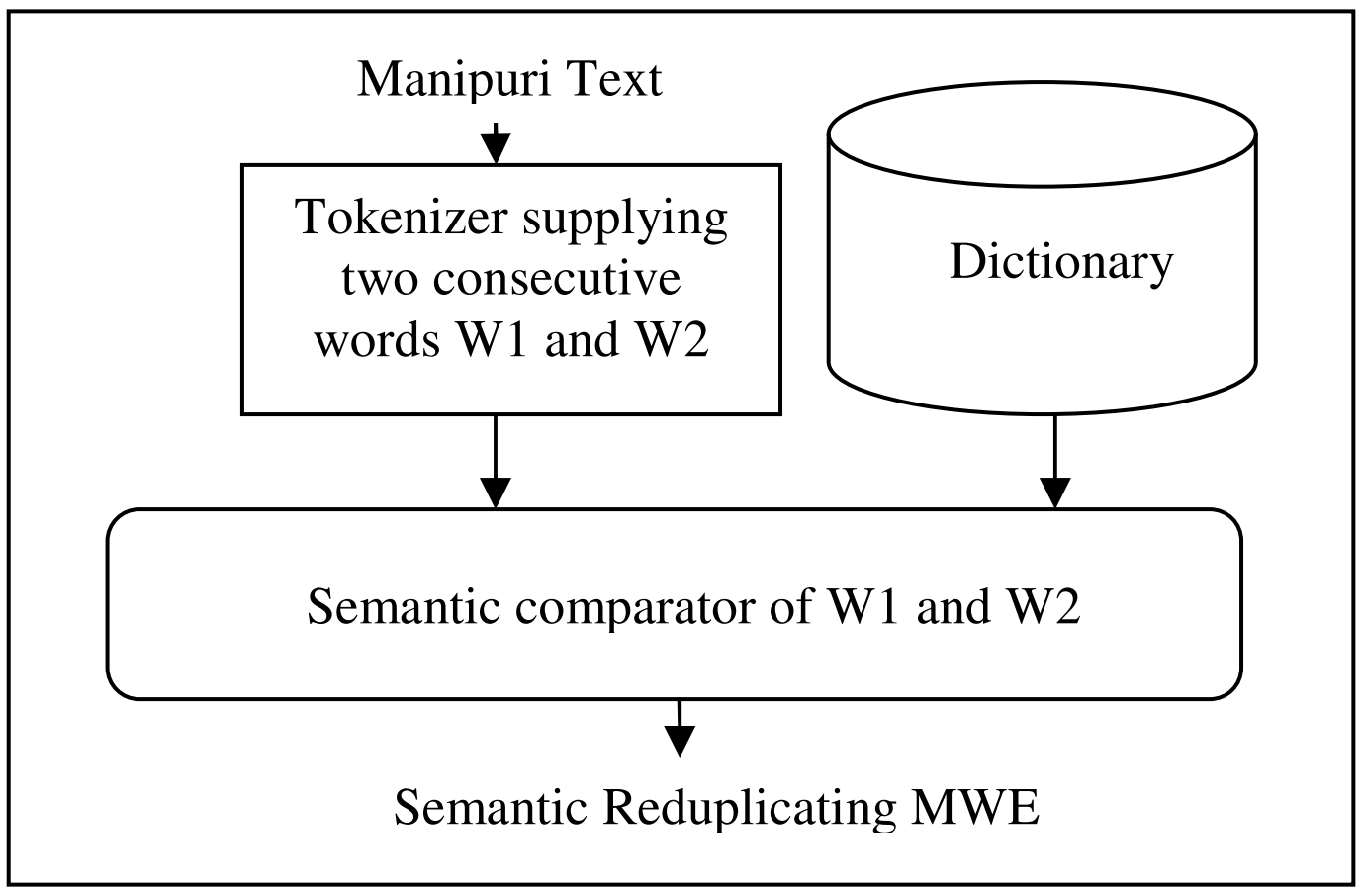

Figure 3: The Second model for identifying semantic reduplicating MWE.

Some of the functions of the second model are same with the first model like **Tokenizer** and **Dictionary** parts but the **Semantic Comparator** works for checking the similarity in semantics of W1 and W2**.**

### 8.1.2 The Algorithm for second model

Algorithm to identify the semantic reduplicating MWEs:

1. *Repeat 2 until all the tokens ($W_i$) are read in the text, where (i=1 to n).*
2. *Check in the dictionary whether $W_i$ and $W_{i+1}$ are semantically same if same identify as* ***semantic reduplication****.*
3. *End.*

## 8.2 The improvement after using RMWE as a feature

After running the training and test files with above model (Section 8.1) the outputs are mark with B-RMWE for the beginning and I-RMWE for the rest of the RMWE and O for the non RMWEs. This output is placed as a new column in the multiple token file for both training and testing.

The training file is run again with the CRF toolkit which outputs a new model file. This model file is used to run the test file which adds up a new output column which is the POS tag by the machine after learning process. This new tagging is used to compare with the previous output. The output shows an improvement of the following:

Table 5: Results RMWEs as a feature in CRF system.

| Model | Recall | Precision | F-Score |
|---|---|---|---|
| CRF | **80.20** | **74.31** | **77.14** |

## 9. CONCLUSION

The previous reported model of [19] it has an accuracy of 72.04% when tested in the test set with CRF system and has an accuracy of 74.38% when tested in SVM system but this purpose system is a better model. This model outperformed the previous model in terms of efficiency with the Recall (**R**) of **78.22%**, Precision (**P**) of **73.15%** and F-measure (**F**) of **75.60%**. With the RMWE as a feature it even betters the efficiency by the Recall (**R**) of **80.20%**, Precision (**P**) of **74.31%** and F-measure (**F**) of **77.14%**.

Of course features selections were manual so a better approach is acceptable to avoid the hit and trial method. A better approach must be the future research area road map because until and unless we have an efficient POS tagger much of the NLP works of Manipuri will be hampered.

## REFERENCES

[1] Nonigopal Singh, N, (1987) *A Meitei Grammar of Roots and Affixes, A Thesis*, Unpublish, Manipur University, Imphal

[2] Thoudam, P. C. *Problems in the Analysis of Manipuri Language*.http://www.ciil-ebooks.net

[3] Brill, Eric., (1992) *A Simple Rule-based Part of Speech Tagger*. In the Proceedings of Third International Conference on Applied Natural Language Processing, ACL, Trento, Italy.

[4] Brill, Eric., (1995) *Transformation-Based Error-Driven Learning and Natural Language Processing: A Case Study in Part of Speech Tagging*, Computational Linguistics, Vol. 21(4), pp543-545.

[5] Black E., Jelinek F., Lafferty J., Mercer R. & Roukos S., (1992) *Decision tree models applied to labeling of texts with parts of speech,* In the DARPA Workshop on Speech and Natural Language, San Mateo, CA, Morgan Kaufman.

[6] Cutting, D., J. Kupiec, J. Pederson & P. Sibun, (1992) *A Practical Part of Speech Tagger*, In the Proceedings of the 3rd ANLP Conference, pp133-140.

[7] Ratnaparakhi, A., (1996) *A maximum entropy Parts- of- Speech Tagger,* In the Proceedings EMNLP 1, ACL, pp133-142.

[8] Kupiec, R., (1992) *Part-of-speech tagging using a Hidden Markov Model*, In Computer Speech and Language, Vol 6, No 3, pp225-242.

[9] Lin, Y.C. T.H. Chiang & K.Y. Su, (1992) *Discrimination oriented probabilistic tagging*, In the Proceedings of ROCLING V, pp87-96.

[10] C. H. Chang & C. D. Chen, (1993) *HMM-based Part-of-Speech Tagging for Chinese Corpora,* In the Proceedings of the Workshop on Very Large Corpora, Columbus, Ohio, pp40-47.

[11] C. J. Chen, M. H. Bai, & K. J. Chen, (1997) *Category Guessing for Chinese Unknown Words,* In the Proceedings of NLPRS97, Phuket, Thailand, pp35-40.

[12] K. T. Lua, (1996) *Part of Speech Tagging of Chinese Sentences Using Genetic Algorithm,* In the Proceedings of ICCC96, National University of Singapore, pp45-49.

[13] Ekbal, Asif, Mondal, S & Sivaji, B., (2007) *POS Tagging using HMM and Rule-based Chunking,* In the Proceedings of SPSAL2007, IJCAI, India, pp25-28.

[14] Ekbal, Asif, R. Haque & & Sivaji, B., (2007) *Bengali Part of Speech Tagging using Conditional Random Field,* In the Proceedings 7th SNLP, Thailand.

[15] Ekbal, Asif, Haque, R. & & Sivaji, B., (2008) *Maximum Entropy based Bengali Part of Speech Tagging,* In A. Gelbukh *(Ed.)*, Advances in Natural Language Processing and Applications, Research in Computing Science (RCS) Journal, Vol.(33), pp67-78.

[16] Smriti Singh, Kuhoo Gupta, Manish Shrivastava, & Pushpak Bhattacharya, (2006) *Morphological Richness offsets Resource Demand –Experiences in constructing a POS tagger for Hindi*, In the Proceedings of COLING- ACL, Sydney, Australia

[17] Sha, F. & Pereira, F., (2003) *Shallow Parsing with Conditional Random fields,* In the Proceedings of NAACL-HLT, Canada, pp134-141.

[18] Doren Singh, T. & & Sivaji, B., (2008) *Morphology Driven Manipuri POS Tagger*, In the Proceeding of IJCNLP NLPLPL 2008, IIIT Hyderabad, pp91-97.

[19] Doren Singh, T., Ekbal, A. & Sivaji, B. , (2008) *Manipuri POS tagging using CRF and SVM: A language independent approach*, In the proceeding of 6th International conference on Natural Language Processing (ICON -2008), Pune, India, pp 240-245.

[20] Kishorjit, N., & Sivaji, B., (2010) *Identification of Reduplicated MWEs in Manipuri: A Rule based Approached*. In the Proceeding of 23rd International Conference on the Computer Processing of Oriental Languages (ICCPOL-2010), Redwood City, San Francisco, pp49-54.

[21] Kishorjit, N., Dhiraj, L., Bikramjit Singh, N., Mayekleima Chanu, Ng. & Sivaji, B., (2011) *Identification of Reduplicated Multiword Expressions Using CRF*, A. Gelbukh (Ed.):CICLing 2011, LNCS vol.6608, Part I, Berlin, Germany: Springer-Verlag, pp41–51.

[22] Kishorjit, N. & Sivaji, B., (2010) *Identification of MWEs Using CRF in Manipuri and Improvement Using Reduplicated MWEs*, In the Proceedings of 8th International Conference on Natural Language (ICON-2010), IIT Kharagpur, India, pp51-57.

[23] Doren Singh, T. & Sivaji, B., (2010) *Web Based Manipuri Corpus for Multiword NER and Reduplicated MWEs Identification using SVM*, In the Proceedings of the 1st Workshop on South and Southeast Asian Natural Language Processing (WSSANLP), the 23rd International Conference on Computational Linguistics (COLING), Beijing, pp35–42.

[24] Lafferty, J., McCallum, A., & Pereira, F. (2001) *Conditional Random Fields: Probabilistic Models for Segmenting and Labeling Sequence Data*, In the Procceedings of the 18th International Conference on Machine Learning (ICML01), Williamstown, MA, USA. pp282-289.

[25] Kishorjit, N., Bishworjit, S., Romina, M., Mayekleima Chanu, Ng. & Sivaji, B., (2011) *A Light Weight Manipuri Stemmer*, In the Proceedings of Natioanal Conference on Indian Language Computing (NCILC), Chochin, India.

**Authors**

Kishorjit Nongmeikapam is working as Asst. Professor at Department of Computer Science and Engineering, MIT, Manipur University. He has completed his BE from PSG college of Tech., Coimbatore and has completed his ME from Jadavpur University, Kolkata, India. He is presently doing research in the area of Multiword Expression and its applications. He has so far published 18 papers and presently handling a Transliteration project funded by DST, Govt. of Manipur, India.

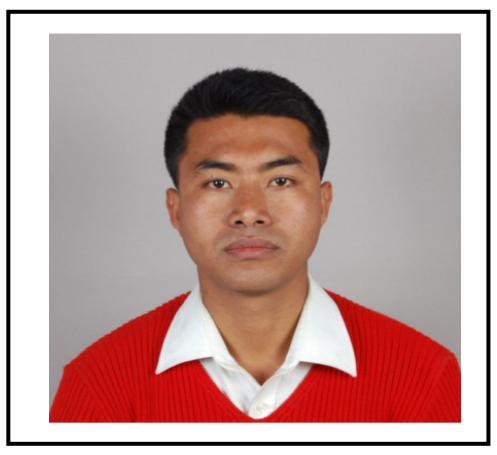

Lairenlakpam Nonglenjaoba is presently a student in MIT, Manipur University. He is pursuing his BE Degree in Computer Science and Engineering. His area of interest is NLP.

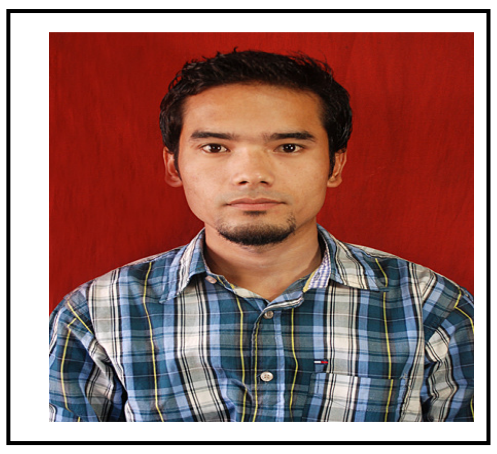

Yumnam Nirmal is presently a student of Manipur University. He is pursuing his MCA in Dept. of Computer Science. His area of interest is NLP.

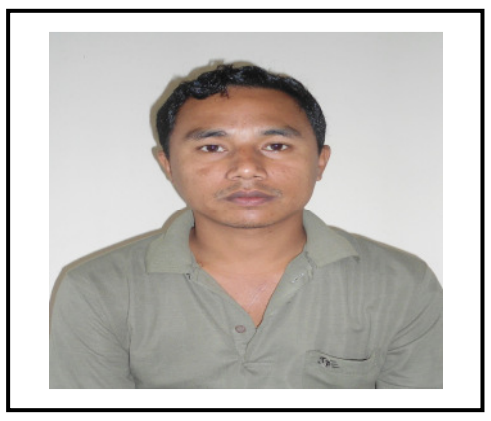

Professor Sivaji Bandyopadhyay is working as a Professor since 2001 in the Computer Science and Engineering Department at Jadavpur University, Kolkata, India. His research interests include machine translation, sentiment analysis, textual entailment, question answering systems and information retrieval among others. He is currently supervising six national and international level projects in various areas of language technology. He has published a large number of journal and conference publications.

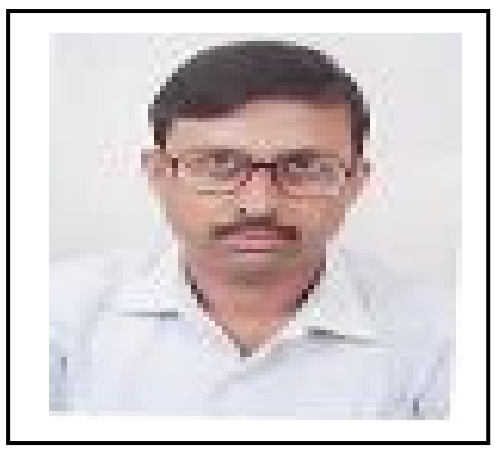